\documentclass[10pt,twocolumn,letterpaper]{article}
\usepackage[top=30mm,bottom=15mm,left=20mm,right=20mm,]{geometry}
\usepackage{times}
\usepackage{epsfig}
\usepackage{graphicx}
\usepackage{amsmath}
\usepackage{amssymb}

\usepackage{url}
\usepackage{amsthm}
\usepackage{algorithm}               
\usepackage{algorithmic}             
\usepackage{multirow}                
\usepackage{bm}
\usepackage{footmisc}

\def\1{\mathbf{1}}
\def\0{\mathbf{0}}

\begin{document}

\title{\LARGE \bf
Retrieving Similar E-Commerce Images Using Deep Learning}

\author{Rishab Sharma$^{1}$,  Anirudha Vishvakarma$^{1}$\\
	$^{1}$Fynd (Shopsense Retail Technologies Pvt. Ltd.)\\ Mumbai, India.\\
	{\tt\small ~~rishabsharma@fynd.com~~anirudhav@fynd.com}\\  \vspace{-1mm}	
}
\maketitle


\begin{abstract}\vspace{-3mm}
In this paper, we propose a deep convolutional neural network for learning the embeddings of images in order to capture the notion of visual similarity. We present a deep siamese architecture that when trained on positive and negative pairs of images learn an embedding that accurately approximates the ranking of images in order of visual similarity notion. We also implement a novel loss calculation method using an angular loss metrics based on the problem’s requirement. The final embedding of the image is combined representation of the lower and top-level embeddings. We used fractional distance matrix to calculate the distance between the learned embeddings in n-dimensional space. In the end, we compare our architecture with other existing deep architecture and go on to demonstrate the superiority of our solution in terms of image retrieval by testing the architecture on four datasets. We also show how our suggested network is better than the other traditional deep CNNs used for capturing fine-grained image similarities by learning an optimum embedding.
\end{abstract}

\vspace{-4mm}
\section{Introduction}\vspace{-1mm}

Finding products that look similar to a particular product is an important feature for a modern e-commerce platform. The visual appearance of a product captures a user’s intent and choices. This information when utilized correctly can boost up a user’s experience and purchase conversions. Collaborative filtering recommends products based on similar user behavior on the platform, but it ignores the product features and also faces the cold-start problem. Retrieval of images using Gabor filters, HOG \cite{c1} and SIFT \cite{c2} are well discovered previously, but are noted to less effective, especially in the case of fashion apparel category, since the performance of these methods largely depend on the representation power of the handcrafted features which are difficult to create. A robust solution here would be the one which can capture fine-grained visual details like shape, pattern, type of print, etc. CNN aids here by converting a product image to an array of numerical embeddings giving the intensity of learned features which differentiates a product. After obtaining this feature vector, a distance matrix can be used to get the visually similar products.
Our approach called RankNet uses a multi-scale siamese network shown in Fig 1. to identify similar images and retrieve them in order of their rank which is a function of the distance between two embeddings in the multi-dimensional space. This distance is calculated using a fractional distance matrix \cite{c3}, unlike the traditional Euclidean distance. Our extensive evaluation has verified that using the fractional distance matrix instead of Euclidean distance not only improves the model accuracy in ranking the images but also in jointly learning the features. Therefore by using supervised similarity information, we can achieve more efficient deep ranking models.
Here, we also address the problems of retrieving a list of visually similar images to a particular query image, both belonging to the same catalog (Visual Recommendations) as well as of retrieving a list of images belonging to the catalog similar to a user-uploaded wild image (Visual Search). The core task addressed by both problems in our work is the quantitative estimation of visual similarity. There are several challenges in dealing with these problems which we have mentioned in this paper. Our image ranking algorithm determines whether a given set of images are visually similar to a particular image by evaluating an image on both higher level and fine-grained visual features. The major progress in image ranking field is in two broad areas: \\
1. Metric learning based \\
2. Image embeddings \\
Each image can be considered a compact feature vector, embedded in a multidimensional space. In recent years many known typical image descriptors like SIFT, HOG and local binary patterns (LBP) \cite{c48} were replaced by some state of the art image CNN which generates feature descriptors. The CNN learns on its own by undergoing supervised training. In order to learn a distance metric, metric-based learning is used which learns from a set of marked training images, plotted in a multi-dimensional embedding space that captures the notion of image similarity. A multi-scale deep Convolutional Neural Network is used by RankNet in form of a Siamese network in order to learn a 4096-dimensional embedding of the query image. In order to project images pairs into a 4096-dimensional space, the network has to learn a set of hierarchical nonlinear transformations during which the network tries to gradually minimize proximity between the positive pair and similarity gradually maximize the proximity for the negative pair. \\
In order to achieve a good model performance and better model convergence, it is really important to choose the right pair of positive images (visually similar images) and negative images (visually non-similar images) for training as a siamese network trains on pairs of images. For consistently fetching the right pair of images, we propose a pair sampling strategy inspired by curriculum learning.
Thus the three major contributions of this paper are:
\vspace{-2mm}
\begin{itemize}
\item A Siamese network consisting of a multi-scale deep CNN that learns a 4096-dimensional embedding of the image to capture the notion of visual similarity. 
\item A fractional distance matrix to calculate the embedding distance between two images in an n-dimensional space instead of the conventional Euclidean distance. 
\item Implementation of an angular loss equation to train a multi-scale CNN to capture fine-grained image similarity between sample images.
\end{itemize}

To determine the performance, we measured the fraction of the correct ordering done by our model. We also compare our proposed RankNet with other state-of-the-art methods for different datasets. The conducted experiments show how RankNet outperforms not only the hand-crafted visual feature-based approaches but also deep ranking models by a considerable margin. \\
We used VGG19 pretrained on Imagenet dataset at the base of our model to get better-initialized weight matrix for training RankNet.

\begin{figure}[htp]
\centering
\includegraphics[width=8cm]{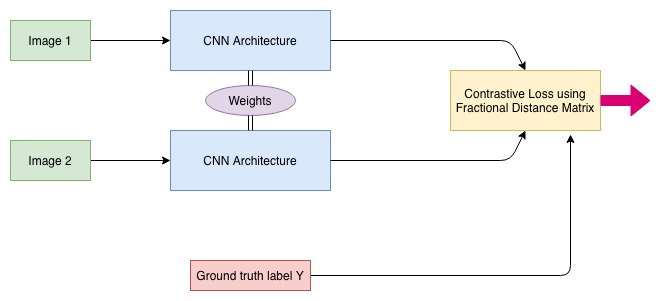}
\caption{Siamese Network Architecture}
\label{fig:captioning}
\end{figure}

\vspace{-2mm}
\section{Related work}
\vspace{-1mm}

A Broad exploration of image similarity has been done using : \\
1. Image content to find similar images \\
2. Text describing the image \\
3. Semantic \\
4. Sketches which help in retrieving similar images \\
5. Approaches based on annotation \cite{c5} \\
All the above approaches use a common computation approach, i.e collect an image database and store it for reference during the inference phase. They aim to compute a similarity function which when given a new image, retrieves similar images from the storage. Earlier image similarity models focused on ways to efficiently crawl and gather reference image data in order to compute similarity. These traditional approaches were little efficient and fast. They use local visual features and other global features [8-11] like color, texture, and shape to design heuristic functions. Some popular ways to compute image similarity were SURF, SIFT, and ORB \cite{c6}. Later in 2005 LeCun et al explored image similarity using convolutional neural networks on a task to retrieve/recognize Handwritten data by using Siamese architecture \cite{c7}.\\
In \cite{c12,c13,c14} the authors have studied image similarity making the models learn on traditional computer vision features like SIFT and HOG. However, the expressive power of these computer vision features makes these model limited. Recently researchers who used deep convolutional neural networks for object recognition have reported great success \cite{c15,c16,c17}. In a deep CNN, the convolutional layers learn a representation of the image with an increasing abstraction level. The descriptor vector which the final layer learns from the image is robust to scale variations and other factors such as viewpoint differences, occlusion, and location of entities within an image. However, when it comes to visual similarity, these descriptors are not much useful since visual similarity is a composite function of both high-level and low-level abstract features/details. The lower-level features are learned to be ignored by an object detection network because for a network need not worry about the color/model of the car to detect it, it simply tries to locate a car shaped object (high-level feature) within the image. This shows that object recognition models learn features common to all the samples in the category, overlooking the details or lower level features which are very important for capturing the notion of visual similarity, thus reducing their effectiveness in the use case of similarity estimation.\\
We compare the results of using features learned in AlexNet \cite{c15}, VGG16 \cite{c4} , and VGG19 \cite{c4}in section four (9). Apart from conventional feedforward networks, siamese networks \cite{c7} are also used for visual similarity assessment. Siamese networks use a contrastive loss function to evaluate the input batch and generate a gradient to optimize the network made up of two CNN's with shared weights. The input to a siamese network is a pair of images which are either similar or dissimilar depending on the ground truth label. Although a Siamese network trains to tackle the very problem of visual similarity which we are addressing, its final prediction being binary (similar/dissimilar) fails the objective of capturing fine-grained visual similarity. Therefore in our approach RankNet, we tackle this architectural bottleneck of binary classification and modify it to learn the fine-grained similarity with the help of densely connected embedding layers.

\vspace{-2mm}
\section{Data Used}
\vspace{-1mm}

In this research, we explored four datasets to train and test our model. Although the performance of our model is evaluated only on Exact Street2Shop dataset.

1. \textbf{Fashion-MNIST} \cite{c18} is Zalando's article image dataset. It has a training set of 60,000 samples and a test set of 10,000 samples. All the sample images of Fashion-MNIST are 28x28 in size and grayscale. All the samples of the dataset belong to 10 object classes namely - top, trouser, pullover, dress, coat, sandal, shirt, sneaker, bag, and ankle boot.

2. \textbf{CIFAR10} \cite{c19} is the second dataset we chose for training RankNet. CIFAR10 is an established dataset used in computer vision for object detection. It has a training set of 50,000 samples and a test set of 10,000 images. All the image samples in CIFAR10 are three channeled colored images and 32x32 in size. Each class of CIFAR10 contains exactly 6000 samples and the test set consists of 1000 randomly-selected images per class. The training images are randomly ordered and each batch of training data contains exactly 5000 samples per class. As CIFAR10 is a publicly available dataset, the class distribution of the dataset is ensured in the train and test subsets.

3. \textbf{Exact Steet2Shop} \cite{c20} is the third dataset we used. The street2shop dataset contains 20,000 images under the wild subset (street images) and 4,00,000 images under the catalog subset (shop images). These photos are categorized across eleven fashion categories that have 39,000 pairs of exact matching products between the shop and the street.

4. The fourth dataset used was published by the authors of the paper \cite{c21} in June 2014. The images in the dataset are hand labeled high-quality triplets. The positive(“p”) and negative(“n”) images belong to the same query text as the query(“q”) image. This dataset provides ranking information for similar images belonging to the same text query. The dataset contains 1599 images which group up to become 5033 triplet pairs. We don't have all the 14000 triplets which are stated in the paper \cite{c21} because the publisher of the data cannot retrieve the public URL links for those images.

The model hyper-parameter and training was optimized by using five-fold cross-validation and the final trained models are evaluated on the basis of accuracy and recall. The test set is only used once at end of the training phase to record the performance of the final model.

\vspace{-2mm}
\section{Architecture}\vspace{-1mm}

\subsection{Deep Ranking Siamese Network}
We see the problem as one of image visual similarity and chose to use a Siamese architecture for learning the embeddings of the data. The siamese network consists of two convolutional neural networks with shared weights which are optimized during training by minimizing a loss function. In our approach, a siamese network is treated as a function f which estimates a particular embedding position for an image I by mapping the sample into an embedding space. The x position in the embedding space can be stated as, given certain parameter $\theta$,  x = f(I;$\theta$) , where I stands for the input of the network (image) and $\theta$ denotes the vector representing all the parameters of the neural network which contains all the biases and neuron weights for the convolution layers as well as the inner product layers. The number of parameters is typically in a range of one million to one-fifty million depending on the architecture and size of the feed-forward network. The aim of the experiment is to produce an embedding with desirable properties by solving for the $\theta$ parameter through the function f such that it places similar images together and dissimilar images apart. The network takes two images as input (see Fig.1) i.e. consider an input pair that contains two different visual views of the same image made using data augmentation or two visual variations within the same apparel category, such a pair is called a positive pair (I\textsubscript{q}, I\textsubscript{p}) and another pair which contain images from different categories, such a pair is called a negative pair (I\textsubscript{q}, I\textsubscript{n}). The input images are then mapped by the network into an embedding space. If the input is (I\textsubscript{q}, I\textsubscript{p}, I\textsubscript{n}) then we can assume the embedding positions to be (X\textsubscript{q}, X\textsubscript{p}, X\textsubscript{n}) and if (X\textsubscript{q}, X\textsubscript{p}) area nearby while (X\textsubscript{q}, X\textsubscript{n}) are further apart then the network has learned a good embedding. A deep convolutional neural network generates these image embeddings. There are Y layers in a deep CNN and Z\textsubscript{y} neurons in the y\textsuperscript{th} layer, where y = 1, 2, 3, ..., Y. When an image x is fed to the network, the y\textsuperscript{th} layer processes it to give an output of the form H\textsubscript{y} = S ( W\textsubscript{y} . x +  B\textsubscript{y} ) where B\textsubscript{y} denotes the bias vector and W\textsubscript{y} denotes the weights of a projection matrix to be learned by the y\textsuperscript{th} layer. s denotes a non-linear activation function here rectified linear unit \cite{c22}. Ultimately we get a non-linear polynomial parametric method f that accurately maps an i dimensional input image to an embedding subspace of e dimensions in the y\textsuperscript{th} layer. The property of this subspace is that it keeps similar images together and dissimilar images further apart. The loss of the network is formalized using the contrastive loss function  \cite{c23} which is employed to the parameters of a parameterized function in such a way that the neighbors are embedded together and the non-neighbors are pushed apart.

\begin{equation}
 L(\theta) = \frac{( 1 - Y)}{2}D(X_q,X_p)^2 + \frac{Y}{2}(max(0,m - D(X_q,X_n)^2)
\end{equation}
\textbf{Eq.1}:  L is the contrastive loss function. The equation calculates loss per sample during training. Any change in m has no affect on the learning as distance matrix simply scales accordingly, m = 1.\\

Eq. 1 defines the training loss for one training pair, where m = 1. It is logically clear that changing the value of m would not impact the learning of the network as the distance metric would simply scale accordingly. In the loss equation, label Y = 1 is assigned to dissimilar or negative image pairs whereas Y = 0 is alloted to similar or positive image pairs. Lastly the deep Convolutional neural networks of the siamese network shares weights which are iteratively optimized using gradient descent by the contrastive loss function L.\\
\begin{figure}[htp]
\centering
\includegraphics[width=8cm]{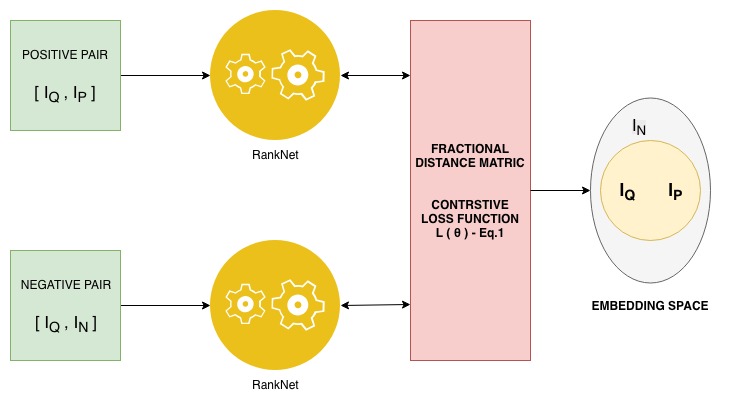}
\caption{In the above illustration positive and negative pairs are fed as input during the training phase. After training, the network tends to embed dissimilar class/category images further as compared to similar/positive images in the 4096 dimensional embedding subspace}
\label{fig:captioning}
\end{figure}
\vspace{-2mm}
\subsection{Contrastive loss function}
Contrastive loss function (L) \cite{c23} is a distance-based Loss function as opposed to prediction error-based loss functions like logistic loss or hinge loss used in classification. Like any other distance-based loss function, it tries to ensure that semantically similar examples are embedded close together. It is calculated on pairs (other popular distance-based Loss functions are Triplet \& Centre Loss, calculated on triplets and point-wise respectively).\\
When similar image pair (label Y = 0) is fed to the network, the right-hand additive section of Eq.1 nullifies and the loss becomes equal to the part containing the positive pair distance between the embeddings of two similar images. Thus if two images are visually similar, the gradient descent reduces the distance between them which is learned by the network. On the other hand, when two dissimilar images(label Y = 1) are fed to the network, the left-hand additive section goes away and the remaining additive section of the equation basically works as a hinge loss function. If the image pair is completely dissimilar and the network outputs a pair of embedding whose proximity is greater than m, then the value of the loss function is maximized to zero else if the images are somewhat similar then we trigger the proximity minimization by optimizing the weights as there is an error. The value m is the margin of seperation between negative and positive samples and is decided empirically. When m is large, it pushes dissimilar and similar images further apart thus acting as a margin. In our work, we have used m = 1.

\subsection{Fractional Distance matrix}

To compute the distance between two embeddings, we used a fractional distance matrix. It has been observed that the Manhattan distance metric provides the best discrimination in high dimensional data spaces. The curse of high dimensionality has a great effect on problems such as nearest neighbor search, indexing, and clustering because in higher dimensional spaces the data metrics become sparser, and the conventional algorithmic and indexing methods fail from an efficiency perspective. The basic concept of proximity or distance is no longer qualitatively meaningful. It has also been observed that under certain reasonable assumptions on the distribution of data, the ratio of the distances of the nearest and farthest point from a given referential point in a high dimensional space approaches 1 for various distance functions and data distributions. Thus in such a case, the problem of nearest neighbor becomes ill-defined as the contrast which distinct two different data points does not exist. Thus in our research, we view the dimensionality curse from an angle of the distance metrics which are used to evaluate the similarity between different subjects. We specifically focused on the use of L\textsubscript{k} norm and inferred that in higher dimensions, the qualitative meaning of proximity is sensitive to the value of k in an L\textsubscript{k} norm. This motivated us to use distance metrics where the value of k is less than 1 ( here 0.2 to 0.3). As stated in \cite{c3}, we will call this distance metrics as fractional distance metrics.

\subsection{Angular Loss}
We also did a lot of experimentation with the loss function employed in our architecture and came to another optimum loss calculation method for training RankNet. In this section we would explain the reason for considering a second loss calculation function for RankNet, which also showed a comparative performance is optimizing the neural network weight matrix.\\
In recent years many forms of deep learning metric have been introduced, but still the major focus of all these forms are either minimizing the Contrastive loss in a Siamese network or the hinge loss in a triplet network. However, it is clearly observed that directly optimizing a distance oriented objective in machine learning is not easy and requires the application of many practical tricks such as hard negative mining and multi-task learning. Recently some work on the N-pair loss and the lifted structure has proposed a better strategy for effectively mining the relation within a mini-batch. But all these studies and works revolves around a common distance based learning between negative and positive pair of images. In our work, we hypothesize that for effectively training deep metric based learning, we must overcome these difficulty by redefining and solving the limitation for the objective in terms of distance. The first limitation of using distance metric is that it is sensitive to scale change. Other than this, it is also noted that the gap between dissimilar clusters is constrained by using the traditional triplet loss methodology. In the above mentioned techniques, different clusters in different scales of intra-class variation are assigned a same absolute margin value, which is logically inappropriate. Also a sub-optimal convergence is achieved in a high-order solution space when we optimize distance-based objectives using stochastic training.\\
In order resolve these issues, we implement a novel angular loss equation proposed by Baidu researchers in \cite{c49} to enhance conventional distance metric learning. The approach is to include the angle formed on the negative anchor by encoding it as a representation of third-order relation existing inside the triplets. Our implemented method pushes the negative sample away from the centroid of the positive cluster by constraining the upper bound of the angle and similarity drags the positive samples towards the centre as the training progresses. The idea behind this implementation is similar to the utilization of high-order details for augmenting pair-wise compulsions in the field of Markov random fields and graph matching. Therefore the implemented angular loss enhances the traditional distance-based loss metric in two ways. Firstly its rotational-invariant and scale-invariant by nature unlike the conventional distance-based metric. This makes our objective of replacing distance-based metric more robust and invariant of the local feature map.\\
However, constraining the angle at the anchor between the the negative and positive sample is more reasonable as it is proportional to the relative ratio among the proximity calculated between the embeddings. Also the angle defines this third-order triangulation within the three samples embedded in the multi-dimensional space. Therefore, given a triplet, the angular loss encodes the local structure of the triplet more accurately than the distance-based triplet loss. Our implementation is broad and can be potentially merged with any other metric learning framework also.  In this version of the paper we have not published the results of angular loss implementation.
\begin{equation}
 L_{A}(\theta) = max(0, D(X_a,X_p)^2 - 4 tan^2 \alpha D(X_a,X_c)^2)
\end{equation}
\textbf{Eq.2}:  L is the angular loss function which calculates the loss per sample triplet during the traing. X\textsubscript{a} is the embedding of the query image, X\textsubscript{p} is the positive image embedding, X\textsubscript{c} is the mean embedding of query and positive and X\textsubscript{n} is the negative image embedding. Alpha is the angle between the negative and positive embedding with query as the anchor point.\\

\vspace{-2mm}
\section{Training Data}\vspace{-1mm}

The training data which is used to train RankNet consists of two type of pair of images - (1) Positive Pair ( Similar images ) and (2) Negative Pair (Dissimilar Images). Before the training, in order to generate these pairs, a query image is randomly sampled from the dataset. After sampling the query image, a set of positive candidate images are programmatically selected from the datset in a bootstrapping fashion with the help of some image similarity scoring techniques. A BISS or basic image similarity scorer need not to be highly accurate or good at recalling. It need not have a good precision ( gets only the similar images), rather it should identify and retrieve the most reasonably alike images that are visually similar to the query image. Thus a basic image similarity scorer should focus on a sub-aspect of visual similarities like color or pattern. Our scorer programmatically selects 100 nearest neighbors from the same class to the query image from the dataset, and create a sample space of the positive images from the retrieved data points. Similarly, negative images are also sampled programmatically by the scorer into two groups, out-of-class samples and in-class samples. The former refers to samples from the same category as that of the query image whereas the latter refers to the samples from some other category as that of the query image. In-class samples teach fine-grained image similarity \cite{c21} to the network as they are not very different from the query image. On the other hand, out-of-class samples teach coarse distinction to the network as they are very different from the query image. In our research, we retrieved the in-class samples from the set union of all the basic image similarity scorers whereas the out-of-class samples were retrieved from the remaining data distribution but within the sample query category / class set. The final sampling was biased so that the data slices contains a ratio of 3:7, which means 30 percent in-class negative samples and 70 percent out-of-class negative samples. We used multiple basic image similarity scorers like ColorHist, AlexNet, and PatternNet. In AlexNet and PatterNet, basic pre-trained model are used  and the FC7 layer features of these network are extracted for encoding the image into a multi-dimensional vector. The foreground of the image is segmented in ColorHist (LAB histogram),  and then after that, the skin is removed.\\
To show the superiority of our methodology, we also explored and compare our approach to a naïve approach of using curriculum learning where the positive and negative images pairs are sampled randomly.

\vspace{-2mm}
\section{Multi-scale Convolutional Neural Network}\vspace{-1mm}

Our aim was to have a high-precision embedding of the images, therefore we used a deep convolutional neural network that incorporates different levels of invariance at various scales \cite{c26,c27}. Deep CNN easily learns to encode strong invariance into their architecture during training, which makes them achieve a good performance for image classification. The strong invariance encoded in the CNN generally grows higher towards the top layers but this growing invariance makes it hard to learn the fine-grained image visual similarity. The final embedding of the image might not be able to capture the simpler sub-aspects of the data sample like colors, pattern and shape. The architecture is shown in Fig. 3 comprises mainly of three CNN's, out of which CNN1 has an architecture similar to that of VGG19's convolutional neural network \cite{c4}. This CNN is used to encode strong invariance and capture the semantics present in the image because it has 19 convolutional layers. Among the 19 layers, the top layers are good at encoding complex representation of image features. The other two CNN's (CNN2 and CNN3 respectively) use a shallower network architecture to capture the down-sampled images. Due to the shallower architecture, these CNN's have less invariance and are used to capture simpler aspects like shapes, pattern, and color which makes the visual appearance of an image. Thus employing three different convolution neural networks instead of a single CNN and making them share lower level layers, makes each CNN independent of the other two. At last, the embeddings from the three convolutional neural networks are normalized and combined with a 4096-dimensional linear embedding layer which encodes and represents an input image as a 4096-dimensional vector. In order to prevent overfitting, we used L2 normalization. Final results show that our multi-scale convolutional neural network outperforms single scale convolutional neural networks on the image similarity task. A major factor responsible for the result is that we combined the embeddings across multiple sub-spaces. The VGG19 \cite{c4} like CNN has a high entropic capacity because of its 4096-dimensional final layer which allows the network to effectively encode the information into the subspaces. Whereas the shallower networks(CNN1 and CNN2) emphasis on fewer dimensions(512 and 1024 dimensions respectively) due to the sparsity of the higher dimensional subspaces.

\begin{figure}[htp]
\centering
\includegraphics[width=8cm]{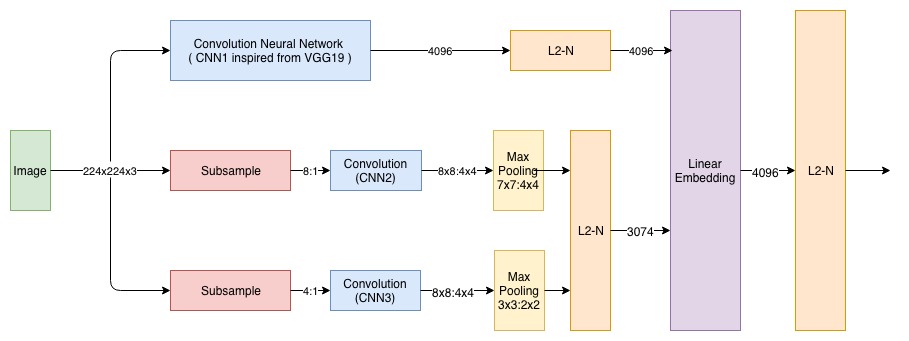}
\caption{Multi-scale convolutional neural network}
\label{fig:captioning}
\end{figure}

\vspace{-2mm}
\section{Implementation}\vspace{-1mm}

Here we shortly demonstrate our observations and implementation details for training a multi-scale CCN like RankNet end-to-end. RankNet is a complex system and our main concerns were preventing and detecting overfitting of the model, especially when employing our training data generation strategy where we do not expose all the possible pairs to the model.
\vspace{-1mm}
\subsection{Deep Ranking Siamese Network}
We fine-tuned pre-trained models using transfer learning \cite{c28}, in order to achieve a faster model convergence. We used a VGG19 like CNN which was initially pre-trained on the ImageNet dataset and fine-tuned the model using a very slow learning rate and RMSProp optimizer which has an adaptive learning rate \cite{c29} instead of an non-adaptive optimizer like stochastic gradient descent \cite{c27}. This gives us more control over the magnitude of the generated updates. We also experimented with the learning rate, decay rate, and optimizer momentum so that the optimizer can continue to make updates towards the global minimum of the loss when the learning rate starts to shrink to smaller numbers. These factors also prevent the network from getting stuck in local minima. We also kept in mind that while fine-tuning a pre-trained model, the updates of the weights using the calculated gradient should be very minute so as not to completely wreck the pre-learned weights, therefore setting the right learning rate is very crucial for convergence. Learning rate can be defined as the step size by which the gradient is multiplied for the network to update itself during backward propagation. It is also noted that if the training loss does not fall very rapidly at the beginning of the initial epoch, then it is advisable to stop the training and adjust the learning rate accordingly. Due to the presence of noise in the training set, selecting the correct number of epochs for training is very necessary for the neural network to converge without overfitting on data. Also the difference between the validation loss of two adjacent epochs give valuable insights into the training phase of the model during cross-validation. Therefore by employing correct number of training epochs and an optimum learning rate, we observe a decreasing trend in the validation loss. The decreasing trend also shows some minor fluctuations throughout it's cycle. We experimented with the hyperparameters like the number of convolutional filters, stride dimensions, padding parameter, learning rate, optimizer and number of layers to get the best fitting solution. Our training set was preprocessed and the data was shifted to unit mean and normalized to speed up the process of convergence.\\
The architecture was implemented in Keras \cite{c32}. The model training was done on a cluster of a nVidia machine, 1 CPU with 16 cores and 4 NVIDIA Tesla K80 GPUs with 2 x 2496 cores and 12 GB and 4 GB RAM respectively. Each epoch (\textgreater10000 iterations) took roughly 5 hours to complete.
\vspace{-1mm}
\subsection{Overfitting}
We augmented the image data with random transformations so that no image appears twice. This helped RankNet to become more robust and prevented overfitting. We also employed dropout in our architecture to prevent overfitting because dropout not only prevents the learning of a redundant pattern by a layer but also acts analogously to data augmentation. Therefore both image augmentation and dropout help to disrupt any random correlations existing in our dataset. It has also been observed that dropout \cite{c30}  and L1 norm \cite{c31} are essentially equivalent to prevent overfitting. This fact helped us while merging the embeddings across different convolutional sub-spaces in our multiscale neural network. We also monitored the number of layers to be trained in the pre-trained CNN because fine-tuning a pre-trained convolutional network is tricky sometimes and it depends on the volume of data. As fine-tuning a pre-trained model using less amount of data can result in an overfitted model [33, 34], so in our case, we only fine-tuned the top two convolutional layers.
\vspace{-1mm}
\subsection{Testing}

The test split of the data was used only once for testing at the completion of training to generate a generalized performance report for our model. Our test set was populated for hyper-parameter tuning and we used five fold cross-validation for selecting our hyperparameters. The test results are reported in the next section of this paper.

\section{Results \& Conclusion}
We evaluated our models on test sets which were split apart from the complete dataset in the beginning. The test sets contain the same number of categories as that in the training set and also the class distribution was similar to that of the training data. The similar distribution also ensures that a generalized performance of the model is being measured.

\begin{figure}[htp]
\centering
\includegraphics[width=8cm]{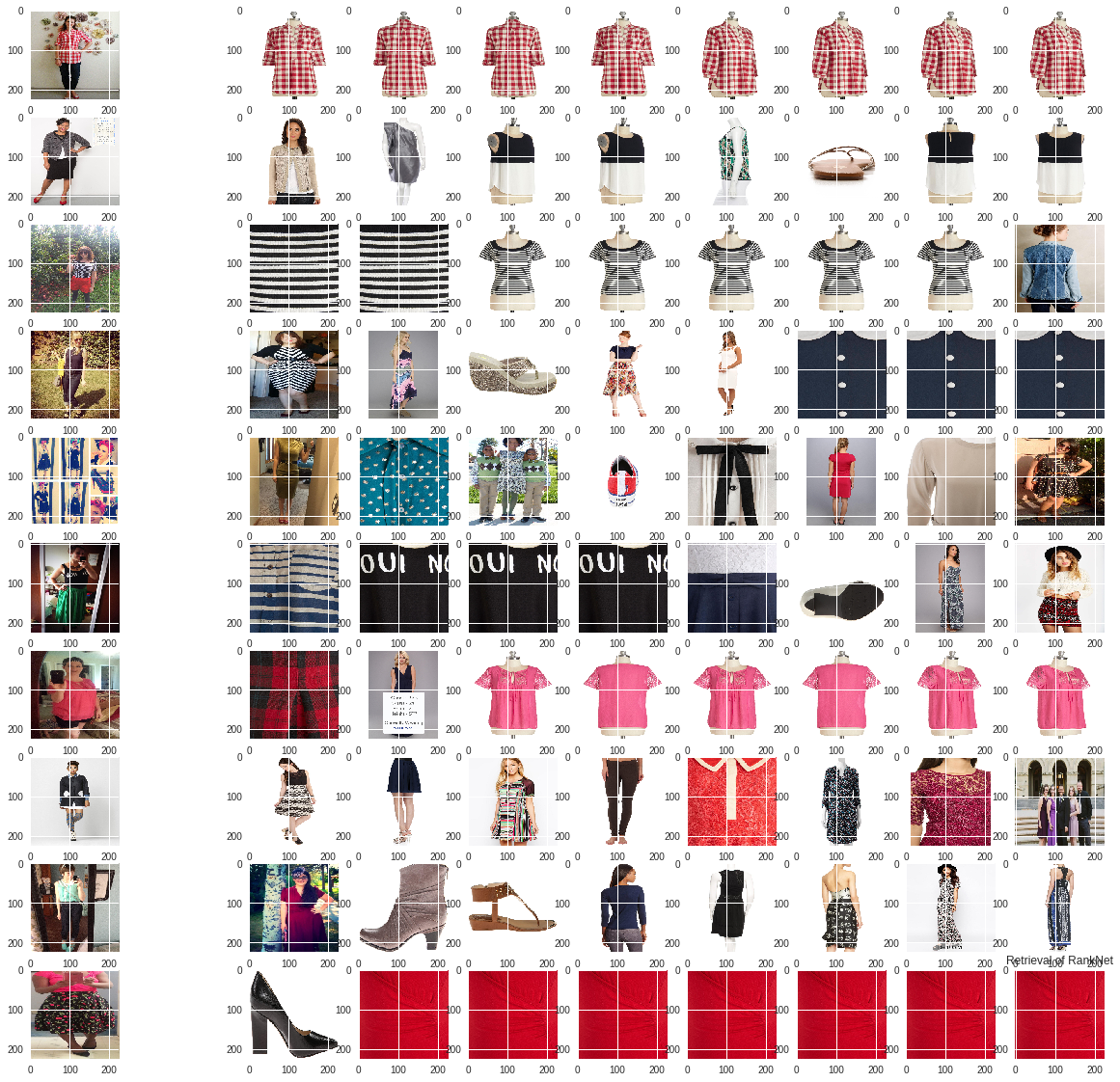}
\caption{Image Retrieval of Ranknet}
\label{fig:captioning}
\end{figure}

\subsection{Embedding space visualization}

In Fig.5 we visualized the embedding space. The embedding space represents the final 4096 dimensions to which an image is mapped. Here we have projected the embedding space to 2D using t-SNE [35]. t-SNE is a distributed stochastic neighbor embedding algorithm used for dimensionality reduction and visualization of high-dimensional datasets. It is implemented using Barnes-Hut approximations which allows it to become applicable on large real-world datasets. 
In our visualization, the different category of images can be seen grouped nearby. Thus showing that in general RankNet performs optimally in projecting similar images close by in the embedding space.

\begin{figure}[htp]
\centering
\includegraphics[width=8cm]{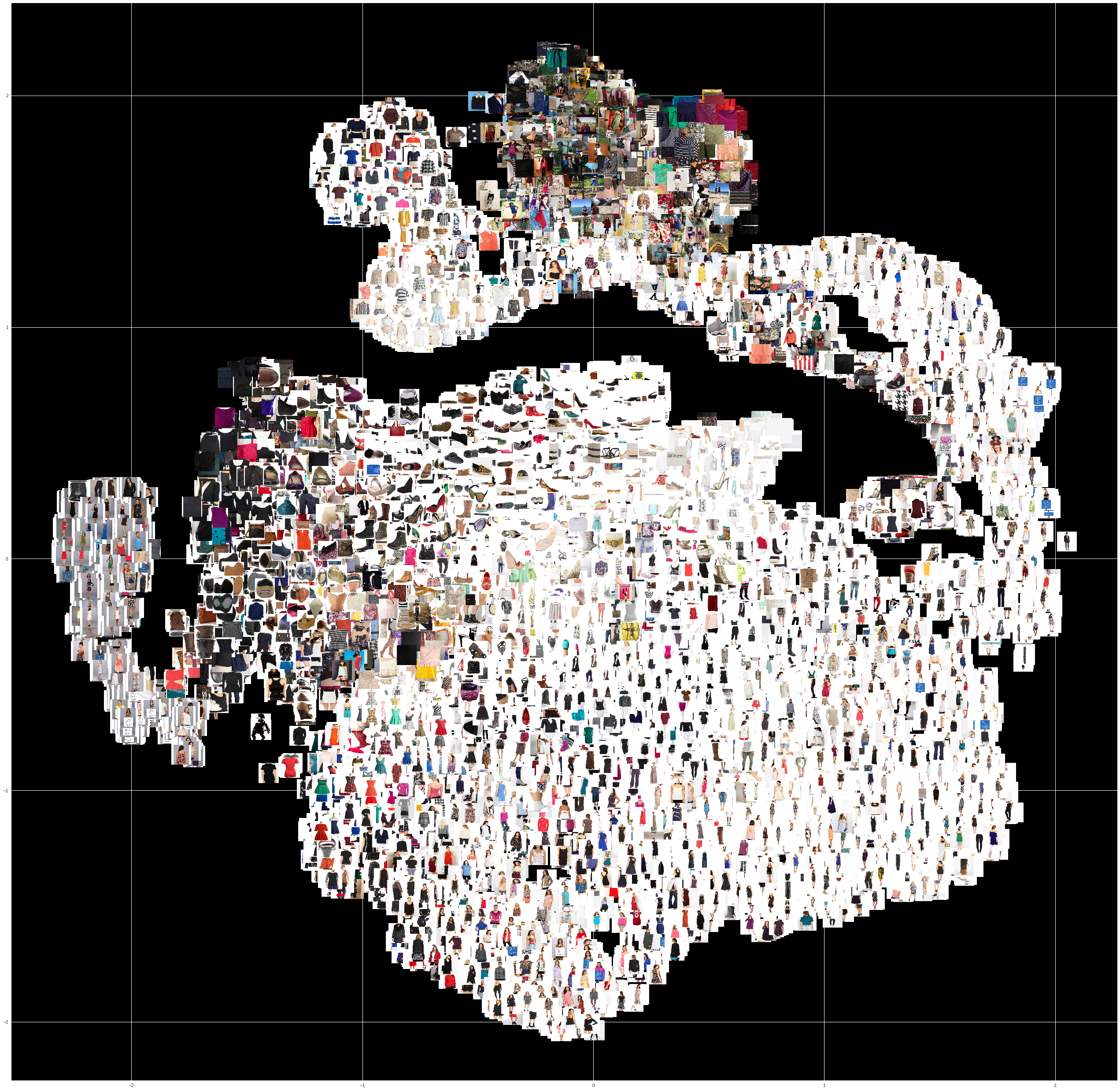}
\caption{t-SNE Visualisation on last layer of RankNet for Street2Shop Test Data - catalog images}
\label{fig:captioning}
\end{figure}

\subsection{Evaluation Metric}

We evaluated all the trained models in terms of accuracy and top-20 recall. The top-20 recall evaluation metric is inspired from \cite{c20}. In top-20 recall it is calculated that in what percent of the cases the correct catalog item matching the query sample in wild image was present in the top-k similar items returned by the model. We employed the contrastive loss function to train the network loss as described in section 4. Baseline model was AlexNet CNN pre-trained on Imagenet. We have shown the evaluaton metric results from the trained models in Table 1 and 2.

\begin{table}[h]
\caption{Validation Accuracy (\%) at Triplet Recall on Exact Street2Shop}
\label{table_example}
\begin{center}
\begin{tabular}{|c||c|}
\hline
Model & Accuracy (\%)\\
\hline
\hline
Ranknet & 94.98\\
\hline
AlexNet & 90.8\\
\hline
Visnet & 93.39\\
\hline
\end{tabular}
\end{center}
\end{table}

\begin{table}[h]
\caption{ Top-20 Recall {\%} on Exact Street2Shop Test Data}
\label{table_example}
\begin{center}
\begin{tabular}{|c||c|}
\hline
Model & Recall (\%)\\
\hline
\hline
Ranknet & 88.576\\
\hline
AlexNet & 14.400\\
\hline
Visnet & 87.914\\
\hline
\end{tabular}
\end{center}
\end{table}

\section{Summary}
In this paper, we presented our architecture known as RankNet to achieve image visual similarity on a given query/reference image. For training, we employed a multi-scale convolutional neural network in a siamese architecture to capture the notion of fine-grained image similarities better than the traditional convolutional neural networks and other deep ranking models which train on triplets \cite{c36} . We also presented a fractional distance matrix to calculate the distance between two data points in a multi-dimensional embedding space, which outperforms the traditional technique of euclidean distance in capturing the idea of proximity in a multi-dimensional space.

\addtolength{\textheight}{-10cm}

\end{document}